\documentclass[sigconf, screen]{acmart}

\usepackage{amsmath}
\usepackage{amsfonts}
\usepackage{graphicx}
\usepackage{xcolor}
\usepackage{soul}
\usepackage{subcaption}
\usepackage{url}
\usepackage{hyperref}

\AtBeginDocument{%
  }

\copyrightyear{2024}
\acmYear{2024}
\setcopyright{rightsretained}
\acmConference[SIGSPATIAL '24]{The 32nd ACM International Conference on
Advances in Geographic Information Systems}{October 29-November 1,
2024}{Atlanta, GA, USA}
\acmBooktitle{The 32nd ACM International Conference on Advances in
Geographic Information Systems (SIGSPATIAL '24), October 29-November 1,
2024, Atlanta, GA, USA}
\acmDOI{10.1145/3678717.3691257}
\acmISBN{979-8-4007-1107-7/24/10}

\newcommand{\model}{\texttt{LM-TAD}}

\begin{document}

\title{Trajectory Anomaly Detection with Language Models}

\author{Jonathan Kabala Mbuya}
\email{jmbuya@gmu.edu}
\orcid{0009-0004-6227-2199}
\affiliation{%
  \institution{George Mason University}
  \streetaddress{4400 University Dr}
  \city{Fairfax}
  \state{Virginia}
  \country{USA}
  \postcode{22030}
}

\author{Dieter Pfoser}
\email{dpfoser@gmu.edu}
\affiliation{%
  \institution{George Mason University}
  \streetaddress{4400 University Dr}
  \city{Fairfax}
  \state{Virginia}
  \country{USA}
  \postcode{22030}
}

\author{Antonios Anastasopoulos}
\email{antonis@gmu.edu}
\affiliation{%
  \institution{George Mason University}
  \streetaddress{4400 University Dr}
  \city{Fairfax}
  \state{Virginia}
  \country{USA}
  \postcode{22030}
}

\begin{abstract}

This paper presents a novel approach for trajectory anomaly detection using an autoregressive causal-attention model, termed \model{}. This method leverages the similarities between language statements and trajectories, both of which consist of ordered elements requiring coherence through external rules and contextual variations. By treating trajectories as sequences of tokens, our model learns the probability distributions over trajectories, enabling the identification of anomalous locations with high precision. We incorporate user-specific tokens to account for individual behavior patterns, enhancing anomaly detection tailored to user context. Our experiments demonstrate the effectiveness of \model{} on both synthetic and real-world datasets. In particular, the model outperforms existing methods on the Pattern of Life (PoL) dataset by detecting user-contextual anomalies and achieves competitive results on the Porto taxi dataset, highlighting its adaptability and robustness. Additionally, we introduce the use of perplexity and surprisal rate metrics for detecting outliers and pinpointing specific anomalous locations within trajectories. The \model{} framework supports various trajectory representations, including GPS coordinates, staypoints, and activity types, proving its versatility in handling diverse trajectory data. Moreover, our approach is well-suited for online trajectory anomaly detection, significantly reducing computational latency by caching key-value states of the attention mechanism, thereby avoiding repeated computations. The code to reproduce experiments in this paper can be found at the following link:~\url{https://github.com/jonathankabala/LMTAD}.

\end{abstract}


\maketitle

\section{Introduction}

\begin{figure}[ht]
\includegraphics[width=.9\columnwidth]{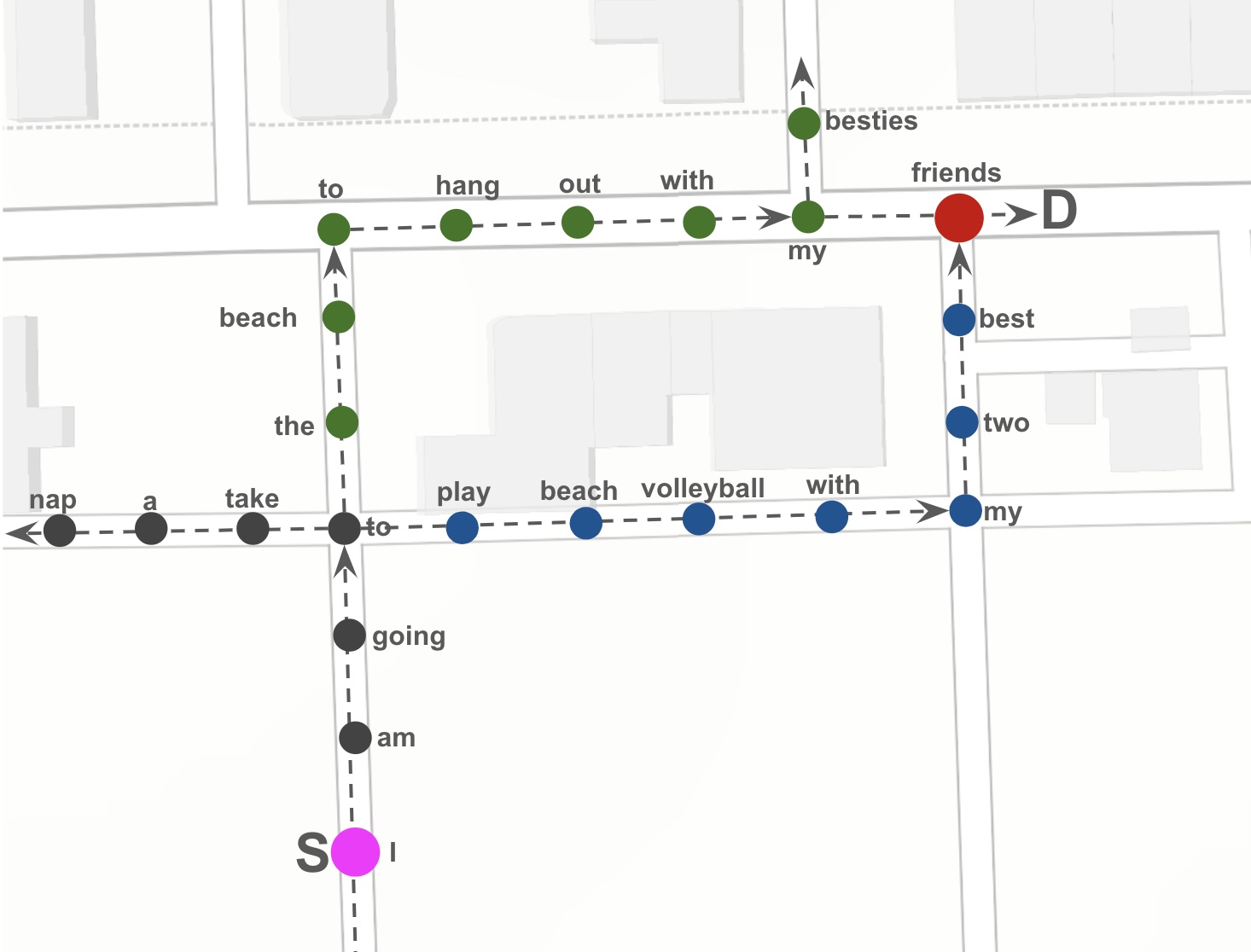}
\caption{A conceptual visualization of trajectories as natural language statements. Language statements and trajectories share similarities: both consist of ordered elements from a finite set (words vs. GPS points) and require connections by semantic or spatiotemporal relationships to be coherent. They are governed by external rules (grammar for the language, road networks for trajectories) and vary by user or context (writing style vs. movement behavior). }
\label{fig:language_as_statment}
\Description{}
\end{figure}  
Effective techniques for gathering and analyzing movement data, including the contribution of this work on anomaly detection are becoming increasingly important with the growth in terms of data and different types of applications.
Specifically, trajectory anomaly detection has several interesting and practical use cases across various fields, such as 
Transportation and Traffic Analysis (accident detection, road safety analysis), Maritime Navigation and Safety (shipping lane monitoring, piracy detection), Air Traffic Control (airspace safety), Wildlife Monitoring (behavior change), Sports Analysis (injury prevention, game strategies), Healthcare and Elderly Care (behavior change and detecting health issues or emergencies), Disaster Response and Management (disaster response and crowd monitoring)
and Urban Planning and Smart Cities (mobility analysis, public transit optimization, pedestrian safety).
This work focuses on detecting trajectory anomalies that deviate from patterns observed in collections of historical datasets. 

Extensive research has been conducted on trajectory anomaly detection for unlabeled data \cite{LiuOnlineAnomalous2020, chec2013iBOAT, Zhang2011IBat, Song2018AnomalyDetectionWithRNN, Kathryn2018IGMM-GANs, Zhang2023OnlineSubtrajectoryDetection}. However, this body of prior work has several limitations.

Firstly, it is difficult to pinpoint specific locations within the trajectory where the anomaly occurs, as the anomaly score is attributed to the entire trajectory or sub-trajectory. Secondly, these methods do not adequately consider anomalies in relation to the individual user's context. This is significant because different users exhibit distinct behavior patterns and what constitutes a normal pattern for one user might be deemed anomalous for another. Finally, anomaly detection has primarily focused on spatiotemporal trajectory data, i.e., GPS coordinates, but the concept of a trajectory can be more abstract. A trajectory can be a chronological sequence of qualitative staypoints, e.g., work $\rightarrow$ restaurant $\rightarrow$ apartment, for a particular user. Additionally, anomalies may not solely relate to the spatial properties of the data, but they can also involve the types of places a user visits, such as a restaurant or a shopping mall on specific days of the week or the duration spent at a particular location.

To address these issues, we propose a \textbf{L}anguage \textbf{M}odel for \textbf{T}rajec\-tory \textbf{A}nomaly \textbf{D}etection, (\textbf{LM-TAD}). The motivation for using a language modeling approach comes from the idea of modeling trajectories as statements \cite{Musleh2022SpeakTrajectories}as illustrated in Figure~\ref{fig:language_as_statment}. Language statements and trajectories share several similarities: 1) both consist of ordered elements from a finite set (words vs. GPS points and segments) and 2) both require coherence, meaning the elements must be connected by semantic (language) or spatial/temporal (trajectories) relationships. For example, in Figure \ref{fig:language_as_statment}, the word \emph{to} cannot be followed by any random word in the vocabulary, just as a GPS point can only be followed by a limited set of other GPS coordinates. 3) Both are governed by external rules—grammar for language and road networks or physical constraints for trajectories. Grammar dictates sentence construction, while road networks dictate possible routes from point A to point B. In Figure~\ref{fig:language_as_statment}, the paths from a source (S) to a destination (D) are constrained by the road network. 4) Finally, the specific combination of elements (words vs. temporal sequence of locations) varies with the user or context. Similar to writing styles, users have different movement behaviors, determining the use of certain words or speed/mobility patterns, respectively. As illustrated in Figure~\ref{fig:language_as_statment}, just as one can choose different words to convey the idea of going to the beach with friends, one can also select different trajectories to travel between a source (S) and destination (D). Based on these similarities, just as a language model can be trained to score the likelihood of sentences, we aim to train a model to score the probability of given trajectories and hence detect anomalous trajectories.

Our specific approach is using an autoregressive causal-attention model to learn the distributions over trajectories. 
We train the model by learning to predict the next location in a trajectory given a historical context. the trained model can compute the probability of generating a location (i.e., discretized GPS coordinate, staypoints, etc.) in a trajectory given its historical context, which in its simplest case is a location history. Anomalies are detected by identifying low-probability locations.
To learn normal behavior for specific users, we can further condition the trajectory generation with a unique user token (i.e., USER\_ID) and flag anomalies on a user basis accordingly. The language model uses discrete tokens and can handle different abstractions of a trajectory, such as discretized GPS coordinates using spatial partitions or qualitative staypoint information (i.e., ``home, work, restaurant, and so forth''). 

To distinguish between normal and anomalous trajectories, we use \emph{perplexity}, a well-established metric in natural language processing. Intuitively, perplexity can be viewed as a measure of
uncertainty when predicting the next token (location) in a
trajectory. We also use the \emph{surprisal rate} of each location to identify anomalous locations to identify the location of an anomaly within a trajectory. 

Our contributions can be summarized as follows: 
\begin{itemize}
    \item We propose a new way to detect anomalies in trajectory data by using an autoregressive causal-attention model. With this approach, we can 1) identify the location in the trajectory where the anomaly occurs, 2) find anomalies with respect to a user, and 3) handle various representations of a trajectory (e.g., GPS coordinates and staypoints)

    \item We show the application of perplexity as a metric for identifying outlier trajectories, both in the context of the entire dataset and with respect to the trajectories of a specific user. Additionally, we illustrate using the surprisal rate to identify potential anomalous locations within a trajectory.
    
    \item Our findings indicate that our method performs exceptionally well on the Pattern of Life dataset (PoL) \cite{ZuflePatternOfLife2023}, effectively identifying anomalous trajectories in the context of a user while training on all data, including anomalies. We also show that our approach is on par with state-of-the-art methods for trajectory anomaly detection when tested on the Porto dataset \cite{YuanTdriveDrivingPorto, YuanDrivingWithKnolwedgePorto} using solely GPS coordinates. Furthermore, our approach is suitable for online anomaly detection as the trajectory is being generated. Unlike autoencoder methods that require the computation of the anomaly score for the entire sub-trajectory each time a new GPS coordinate is sampled, our method benefits from low latency by caching key-value (KV cache) states \cite{Liu2024MiniCacheKC, Pope2022EfficientlyST} of the attention mechanism for previously generated tokens (i.e., GPS coordinates), thereby avoiding repeated computations.
\end{itemize}

The remainder of this paper is organized as follows.  Section \ref{related_work} discusses related work. Section \ref{problem_formulation} gives the basic formulation of the problem. In Section \ref{method},  we present our autoregressive generative approach to detect anomalies in trajectories. Section \ref{experiments} provides an experimental evaluation that highlights the benefits of our method compared to existing approaches. Finally, Section \ref{conclusion} concludes and provides directions for future work.

\section{Related Work} \label{related_work}
\subsection{Trajectory Anomaly Detection}
Existing work for anomaly detection in trajectories can be grouped into two broad categories: heuristic-based methods \cite{Lee2008TRAOD, Zhang2011IBat, chec2013iBOAT, Zhu2015Time_dependent_popular, Zhongjian2017}  and learning-based methods \cite{Song2018AnomalyDetectionWithRNN, Smolyak2020GANsforAnomalyDection}. 

Heuristic-based methods primarily rely on hand-crafted features to represent normal routes and employ distance or density metrics to compare a target route to normal routes. The study in \cite{Lee2008TRAOD} suggests a partition-and-detect framework for trajectory outlier detection, effectively identifying outlying sub-trajectories by combining two-level trajectory partitioning with a hybrid distance-based and density-based detection approach. Studies by \cite{Zhang2011IBat} and  \cite{chec2013iBOAT} introduce related methods that systematically extract, group, and analyze trajectories based on the source and destination. These methods identify anomalies by how rare they are and how much they deviate from usual patterns, using the principle of isolation to ensure effective and reliable anomaly detection. Another research effort by \cite{Zhu2015Time_dependent_popular} presents a time-dependent approach for detecting trajectory anomalies, employing edit distance metrics to ascertain whether a given target trajectory deviates significantly from historical normal trajectories. Similarly, \cite{Zhongjian2017} uses edit distance coupled with a density-based clustering algorithm to identify anomalous trajectories. Heuristic-based methods exhibit certain limitations. Primarily, the characterization of a trajectory is dependent on manually curated features encompassing various parameters, such as frequency, distance, or density thresholds, to flag a trajectory as anomalous. Furthermore, the construction of these features tends to be domain-specific, necessitating specialized expertise in the respective field. Additionally, the applicability of these methods across different regions is constrained owing to the inherent dissimilarities in trajectories across diverse geographical locations.

Learning-based methods rely on machine learning techniques. The work by \cite{Song2018AnomalyDetectionWithRNN} employs trajectory embedding learned by recurrent neural networks (RNN) to capture the sequential information and distinctive characteristics of trajectories to detect anomalies. However, the model requires labeled data, usually unavailable in real applications due to the cost of labeling a dataset. Several studies on unlabelled data have been proposed to overcome the limitations of using labeled data. \cite{Smolyak2020GANsforAnomalyDection} suggest a method combining Infinite Gaussian Mixture Models with bi-directional Generative Adversarial Networks to detect anomalies in trajectory data using a combination of the reconstruction loss and discriminator-based loss. Deep learning methods based on autoencoders have recently been applied to various anomaly detection tasks \cite{Malhotra2016LSTMbasedEF, Zhou2017AnomalyDetectionAE, Jinghui2017AEnsembles, Zong2018DeepAG}. These methods work by learning to compress and reconstruct the input. They use the premise that anomalous input will produce a significant reconstruction error, as they differ from the learned normal patterns. A recent study by \cite{LiuOnlineAnomalous2020} proposes an autoencoder method for online anomalous trajectory detection with multiple Gaussian components in the latent space to discover various types of normal routes. Outside of autoencoder methods, another recent study by \cite{Zhang2023OnlineAnomalousSub} suggests a reinforcement learning-based solution for detecting anomalous trajectories and sub-trajectories. However, the methods have two main limitations. Firstly, these methods are limited in pinpointing the specific location of an anomaly within the trajectory, as they rely on an aggregate anomaly metric, typically the reconstruction error. Secondly, there is a notable lack of generalizability in these approaches to scenarios requiring user-specific anomaly identification, as what constitutes an anomaly for one user might be deemed normal for another.

\subsection{Language Modeling on Trajectory Data}
The field of language modeling has received much attention recently since the introduction of the transformer model \cite{Vaswani2017AttentionIA}. Language models like BERT \cite{devlin-etal-2019-bert}, GPT-2 \cite{Radford2019LanguageMA}, and LLaMA \cite{Touvron2023LLaMAOA} have been shown to achieve great performance on a variety of natural language tasks, including question-answering, sentiment analysis, and text generation. Language modeling techniques have been extended to other applications, including image classification \cite{Radford2021LearningTV, Zhou2021LearningTP} and speech processing \cite{Baevski2020EffectivenessofASR, Baevski2020wav2vec2A}. Recent studies have also applied language modeling techniques to a wide range of applications on mobility data. For example, the work in \cite{Xue2022LeveragingLF} leverages language modeling techniques for human mobility forecasting tasks, while the work in \cite{Hong2022HowDY} uses similar techniques to predict the next visited location in a trajectory. The work in \cite{Musleh2022SpeakTrajectories} proposes a conceptualization of a BERT-inspired system tailored for trajectory analysis. However, none of the previous work used a generative approach for anomaly detection in trajectory data.

\section{Problem Formulation} \label{problem_formulation}
A trajectory is a finite chronological sequence of \emph{visited locations} and can be modeled as a list of space-time points modeled as location and time stamp pairs $T=p_0, \ldots, p_n$ 
with $p_i=\langle l_i, t_i\rangle$ 
and $l_i \in R^2, t_i \in R^+$ 
for $i=0,1, \ldots, n$ and $t_0 < t_1 < t_2 < \ldots < t_n$ (cf. \cite{pfoser1999ssd}).

In the simplest case, a location $l_i$ is represented as a geographic coordinate in two-dimensional space. 
Other representations are to map locations to cells of a discretized space such as a regular spatial or a hexagonal grid \cite{uberh3}. 

Alternatively, $l_i$ can capture qualitative staypoints (visited points of interest such as ``home'', ``work'', or ``restaurant'') or spatial partitions that capture functional areas of a city, e.g., ``commercial'', ``business'', or ``residential'' areas. Therefore, $l_i$ can include both a staypoint and afunctional area, e.g., $l_i =[\text{apartment}, \text{ downtown}]$).

A collection of related trajectories $T_i$ constitutes a dataset $\mathcal{D}$.
The dataset $\mathcal{D}$ may contain both normal and anomalous trajectories. In general, an anomalous trajectory refers to one that does not show a \emph{normal} mobility pattern and deviates from the majority of the trajectories in $\mathcal{D}$ \cite{Zhang2023OnlineSubtrajectoryDetection, Chen2012AnomalyOnTaxi}.
Given a dataset $\mathcal{D}$ with $n$ trajectories, our goal is to train a model that distinguishes between normal and anomalous trajectories without having explicit labels.

\section{Method} \label{method}

\begin{figure}
\includegraphics[width=1 \columnwidth]{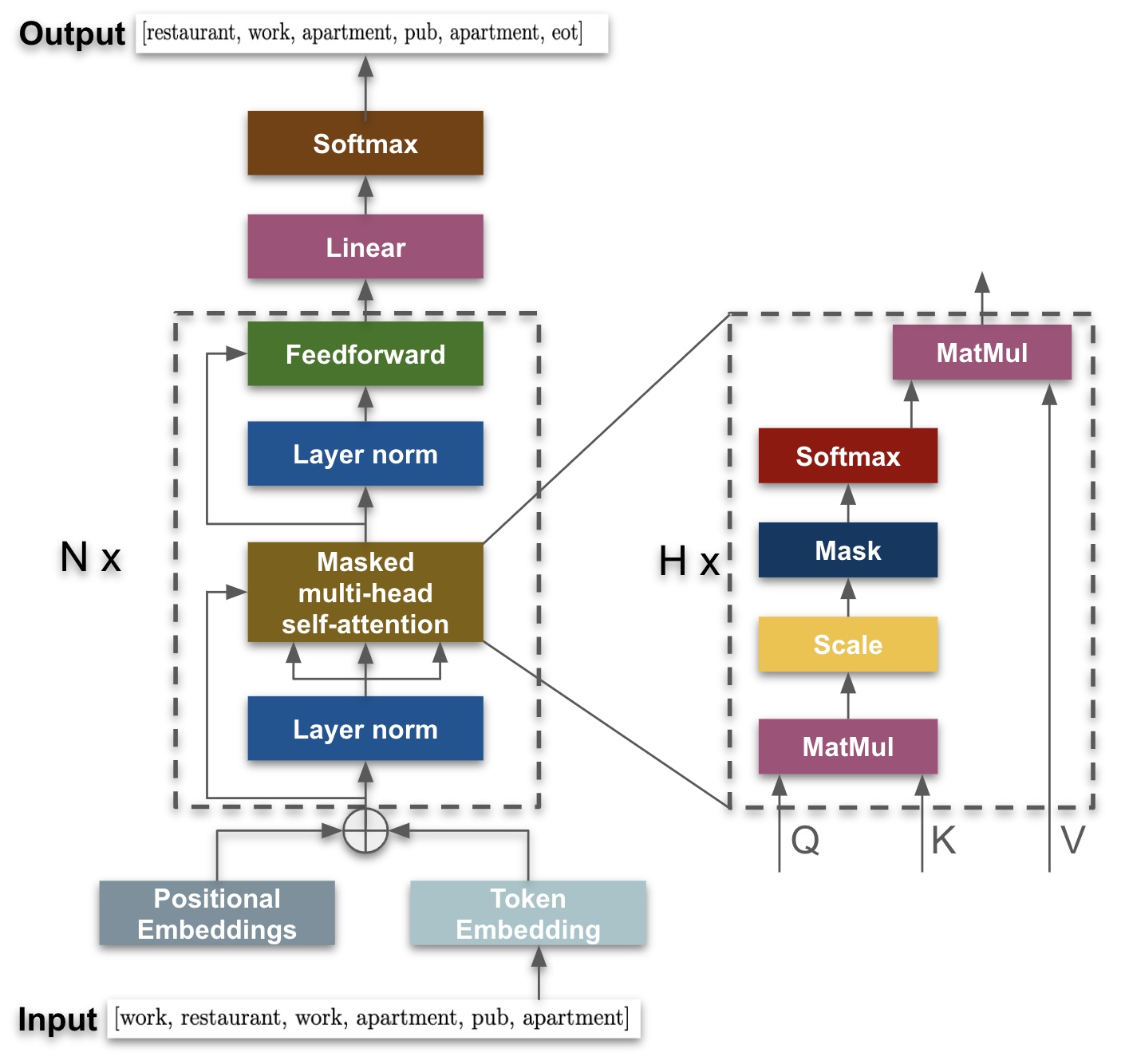}
\caption{Architecture of \textbf{\model{}}, our trajectory model. }
\label{fig:model}
\Description{}
\end{figure}  

\begin{figure*}
     \centering
     \begin{subfigure}[b]{0.45\textwidth}
         \centering
         \includegraphics[width=\textwidth]{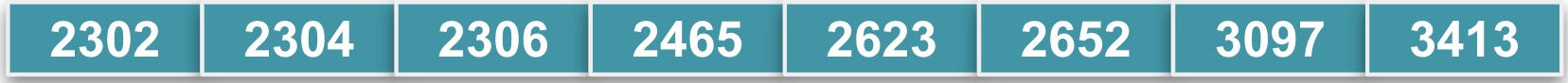}
         \caption{Discretized GPS}
         \label{fig:tokens}
     \end{subfigure}
     \begin{subfigure}[b]{0.45\textwidth}
         \centering
         \includegraphics[width=\textwidth]{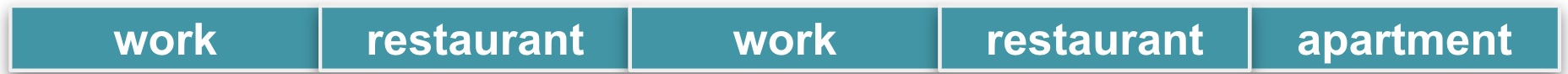}
         \caption{Staypoints}
         \label{fig:staypoints}
     \end{subfigure}
     \hspace{1em}
     \begin{subfigure}[b]{0.45\textwidth}
         \centering
         \includegraphics[width=\textwidth]{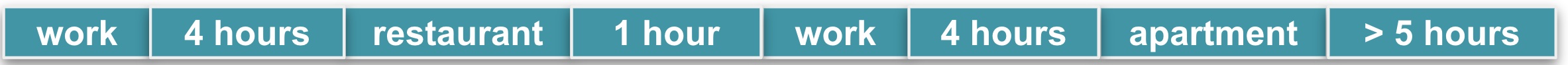}
         \caption{Staypoints + duration}
         \label{fig:staypoints_hours}
     \end{subfigure}
     \begin{subfigure}[b]{0.45\textwidth}
         \centering
         \includegraphics[width=\textwidth]{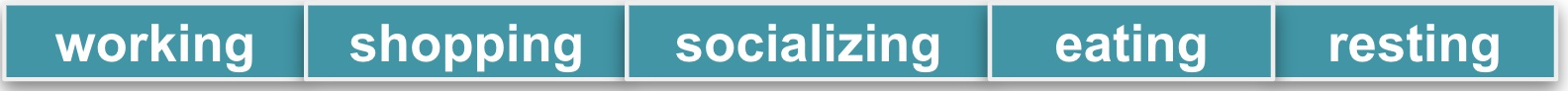}
         \caption{Activities}
         \label{fig:activites}
     \end{subfigure}
        \caption{Example location configurations. Locations can be (a) discretized GPS coordinates, (b) staypoints, (c) staypoints enhanced with dwell time, or (d) activities.   }
        \label{fig:location_configuration}

    \label{fig:method_comparisions}
    \Description{}
\end{figure*}

Our approach is to train a model that learns probability distributions over trajectories. An autoregressive generative model will allow us to infer the probability of a trajectory given the historical context
\begin{equation}
    P(T) = p(l_1)p(l_2|l_1)p(l_3|l_1l_2) \ldots p(l_i|l_{<i}) \ldots p(l_n|l_{<n})
\end{equation}
 
where the probability of each location $p(l_i|l_{<i})$ is conditioned on a complete location history. We note that there is no time bound between locations. However, we could use such information as part of the input. With this approach, we can find anomalous trajectories and identify exactly which locations in the trajectory are anomalous.

\subsection{Model and Architecture}
\noindent
Given a dataset of trajectories $\mathcal{D} = \{T_1, \ldots, T_m\}$, \textbf{\model{}}'s goal is to maximize the likelihood of all the trajectories in the dataset:
\begin{equation}
    \mathcal{L(\mathcal{D})} = \sum_{i=1}^{ |\mathcal{D}|} \sum_{j=1}^{|T_i|} \log P(l_i|l_{<i}; \theta)
\end{equation}
where $P(\cdot ; \theta)$ is the conditional probability modeled by a neural network parameterized by $\theta$. To learn the parameters $\theta$, we opt for a transformer-based network architecture \cite{Vaswani2017AttentionIA}. This architecture choice is motivated by its proven efficacy in natural language generation tasks, suggesting its potential applicability and effectiveness in modeling trajectories as statements.

Figure~\ref{fig:model} shows the overall architecture of our method, \textbf{\model{}}, which consists of positional and token embeddings, $N$ transformer blocks followed by a linear transformation, and a softmax layer. 

To capture input semantics, the token embedding layer transforms each token (location) from a categorical type to a finite-dimensional real-valued vector. Positional embeddings play a critical role in the training process, compensating for the absence of inherent sequential ordering within the causal-attention module. Each transformer block comprises a multi-head causal-attention mechanism, which is preceded and succeeded by a layer-normalization layer and a feedforward layer. In the multi-head causal-attention mechanism, a trajectory is transformed into three sets of vectors —keys, values, and queries— and then split into multiple heads for parallel processing. Each head independently computes a scaled dot-product attention to get attention scores that assess the relevance of different locations (tokens) in a trajectory. This allows the model to concurrently learn dependencies between locations, such as temporal or spatial ones. The outputs from all heads are concatenated and linearly transformed to produce the final output. Additionally, the causal-attention mechanism includes a masking operation to prevent the attention function from accessing information from future tokens (locations), given the autoregressive nature of our approach. 

Below is the formal description of the muti-head self-attention:
\begin{equation}
    \begin{aligned}
        \mbox{Attention}(\mathbf{Q}, \mathbf{K}, \mathbf{V}) = \mbox{softmax} \left(\frac{\mathbf{Q} \mathbf{K}^T}{\sqrt{d_k}} \right)  \mathbf{V}
    \end{aligned}
\end{equation}
where $d_k$ is the dimension of the keys. Concatenating the output values results in:
\begin{equation}
    \begin{aligned}
        \mbox{MultiHead}(\mathbf{Q}, \mathbf{K}, \mathbf{V}) = \mbox{Concat(head$_1$, ..., head$_h$)}   \mathbf{W}^{o} \\
        \mbox{with head}_i = \mbox{Attention}(\mathbf{Q} \mathbf{W}^Q_i, \mathbf{K} \mathbf{W}^K_i, \mathbf{V} \mathbf{W}^V_i)
    \end{aligned}
\end{equation}
where the $\mathbf{W}^Q_i \in \mathbb{R}^{d_{model} \times d_{q}}$,  $\mathbf{W}^K_i \in \mathbb{R}^{d_{model} \times d_{k}}$, $\mathbf{W}^K_i \in \mathbb{R}^{d_{model} \times d_{v}}$ are projection matrices that are learned during training. The projection matrix $\mathbf{W}^{o}$ linearly combines the outputs from different attention heads, enabling the model to flexibly adjust and fine-tune the aggregated attention, thereby enhancing the model's capacity to learn complex patterns. In \textbf{\model{}}, $d_q = d_k = d_v = d_{model}/h$ where $h$ is the number of heads. 

The feedforward layer consists of two linear transformations linked with a ReLU activation function:
\begin{equation}
    \begin{aligned}
        \mbox{FFM}(x) = \mbox{max}(0, \mathbf{x}\mathbf{W}_1 + \mathbf{b}_1) \mathbf{W}_2 + \mathbf{b}_2
    \end{aligned}
\end{equation}
where the weights $\mathbf{W}_1 \in \mathbb{R}^{d_{model} \times d_{ff}}$, $\mathbf{W}_2 \in \mathbb{R}^{d_{ff} \times d_{model}}$ and the biases $\mathbf{b}_1 \in \mathbb{R}_{d_{ff}}$, $\mathbf{b}_2 \in \mathbb{R}_{d_{model}}$. The transformer block uses the layer-normalization layer in addition to residual connections to stabilize learning and improve training efficiency.

The output of the transformer block goes through linear and softmax layers to predict the distribution of each token in a trajectory. 

\subsection{Location Configurations}
An advantage of using a generative approach to model trajectories is the ability to abstract the locations of a trajectory in different ways. Figure~\ref{fig:location_configuration} provides examples. In the simplest case, a trajectory can be represented by a finite chronological sequence of GPS coordinates. These coordinates can be discretized using regular grid cells (Figure~\ref{fig:tokens}) \cite{Li2018DeepRepLearningforTraj}  or hexagons \cite{Fan2022OnlineTrajectoryPrediction}. However, various other trajectory configurations are possible. Instead of GPS coordinates, we can also use staypoints (``home'', ``workplace'', ``restaurant'', etc.)  (Figure~\ref{fig:staypoints}) or points of interest. We can even model a trajectory as a chronological sequence of a person's activities (``eating'', ``working'', and ``playing sports'')  (Figure~\ref{fig:activites}), where each location in the trajectory corresponds to the activity of a person at a particular time. These trajectories can even be enhanced with additional metadata, such as the dwell time at a location (Figure~\ref{fig:staypoints_hours}), method of transportation, or proximity to the previous location. An advantage of \model{} is its ability to work with any of these different trajectory configurations. 

\subsection{Anomaly Score}
We use \emph{perplexity} to determine how anomalous a trajectory is. Perplexity is a well-established measure to evaluate language models \cite{peters-etal-2018-deep,devlin-etal-2019-bert,gpt-3}, and can be viewed as a measure of uncertainty when predicting the next token (location) in a trajectory \cite{Ngo2021NoNI}. Equation~\ref{eq:trajectory-perplexity} shows how we can calculate the perplexity of a trajectory $T$ with $t$ locations, and Equation~\ref{eq:dataset-perplexity} shows how we compute the perplexity over a dataset $\mathcal{D}$ with $n$ trajectories $T_i$. 

\begin{equation}\label{eq:trajectory-perplexity}
    PPL(T) = \exp\left \{ - \frac{1}{n} \sum^t_{i=1} \log P(l_i|l_{<i}) \right \}
\end{equation}

\begin{equation}\label{eq:dataset-perplexity}
    PPL(\mathcal{D}) = - \frac{1}{n} \sum^n_{i=1} PPL(T_i) 
\end{equation}

The \textit{lowest possible perplexity is 1}, which implies that the model can correctly predict the next location with absolute certainty. However, the maximum of this measure is unbounded. To determine when a trajectory is anomalous (``high'' perplexity), we need to provide a threshold. We note that the choice of a threshold can be application- and dataset-dependent \cite{An2015VariationalAB}. We can compute the threshold as follows:
\begin{equation*}
    \text{threshold} = \text{mean}[PPL(\mathcal{D})] + \text{std}[PPL(\mathcal{D})]
\end{equation*}
where $\text{mean}[PPL(\mathcal{D})]$ and $\text{std}[PPL(\mathcal{D})]$ are the mean and standard deviation of the perplexities of $n$ training trajectories.
To identify abnormal trajectories with respect to a specific user, we customize the threshold for each user. Here, the mean and standard deviation will be computed only using the training samples of that user.

\section{Experiments} \label{experiments}

This section presents the experimental setup used to evaluate the effectiveness of our proposed \model{} model. We compare our method to several state-of-the-art baselines using two different datasets and utilizing different evaluation metrics to measure the accuracy and robustness of anomaly detection.

\subsection{Datasets \& Preprocessing}
We use simulated and real-world datasets. Specifically, we use the Pattern-of-Life (PoL) simulation dataset~\cite{ZuflePatternOfLife2023} and the Porto taxi dataset \cite{YuanTdriveDrivingPorto, YuanDrivingWithKnolwedgePorto}.

\subsubsection{Pattern-of-Life Dataset (PoL)} 
\label{pol_description}

The PoL dataset was generated through the Pattern-of-Life (PoL) simulation \cite{ZuflePatternOfLife2023, amiri2023massive}. This simulation consists of virtual agents designed to emulate humans' needs and behavior by performing human-like activities. Activities include going to work, restaurants, and recreational activities with friends. These activities are performed at real locations obtained from OpenStreetMap \cite{OpenStreetMap2023}. While agents engage in these activities, the simulation also records the location, which includes the GPS coordinates and the staypoints (i.e., home, work, restaurant), as well as the respective timestamps.

Using the raw data from the PoL simulation, we created daily trajectories for each agent, consisting of places they visited on that particular day. The geographic coverage was Atlanta, GA and we simulated the behavior of an agent population consisting of working professionals. 

The dataset includes an average of 450 daily trajectories for each of the 1000 generated agents, resulting in a total of 444,634 trajectories. Each input to the model represents a virtual agent's daily trajectory. To capture the patterns of each agent, the agent ID is included at the beginning of the trajectory. Based on the hypothesis that individual behavioral patterns exhibit consistency on the same days of the week, we also incorporate weekday information into the feature vector to enhance the model's ability to detect anomalies.

For example, a daily trajectory is represented as: [agent\_ID, weekday, work, restaurant, apartment]. We also consider other location representations, including discretized GPS coordinates and the duration of stay at a location. We use Uber hexagons \cite{uberh3} for discretized locations and discretize the stay duration into 1-hour buckets using a sequence of bucket IDs as input to the model.

The PoL dataset comes with labels to identify anomalous trajectories generated by the simulation. To introduce anomalies, the simulation selects ten virtual agents exhibiting anomalous behaviors. For example, work anomaly is one type supported by this simulation: agents with work anomalies will abstain from going to work when they typically would. 

For agents with anomalous trajectories, we have the first 450 days representing normal behavior and the last 14 days that exhibit anomalous behavior. 

We trained our model on the entire dataset, including the additional 14 days of anomalous behavior from the ten virtual agents, to ensure it could identify outliers even when they were present in the training data. We then tested our methods against the baselines using the entire dataset

\begin{table*}[t]
\caption{Outlier detection on RED outlier agents for the Pattern of Life dataset. The best results are \textbf{bolded}. \model $\space$ clearly outperforms the baselines, showcasing its modality adaptation capabilities}
\label{tab:p_of_life_results}
\centering
\begin{tabular}{r|cc|cc|cc|cc} \toprule

& \multicolumn{6}{c|}{baselines} &   \multicolumn{2}{c}{ours} \\
  & \multicolumn{2}{c|}{SAE} &   \multicolumn{2}{c|}{VSAE} & \multicolumn{2}{c|}{GM-VSAE} &   \multicolumn{2}{c}{LM-TAD}  \\ \midrule
 
Agent & F1 & PR-AUC & F1 & PR-AUC  & F1 & PR-AUC & F1 & PR-AUC   \\ \midrule
 57 & 0.00 & 0.01 & 0.00 & 0.04  & 0.16 & 0.03 & \textbf{0.72} & \textbf{0.59}   \\ %
 62 & 0.00 & 0.11 & 0.00 & 0.02  & 0.10 & 0.09 & \textbf{0.87} & \textbf{0.85}   \\ %
 347 & 0.00 & 0.02 & 0.00 & 0.02  & 0.25 & 0.40 & \textbf{0.78} & \textbf{0.78}   \\ %
 546 & 0.00 & 0.07 & 0.00 & 0.03  & 0.12 & 0.12 & \textbf{0.75} & \textbf{0.63}   \\ %
 551 & 0.00 & 0.01 & 0.00 & 0.02  & 0.00 & 0.04 & \textbf{0.76} & \textbf{0.63}   \\ %
 554 & 0.00 & 0.02 & 0.00 & 0.02  & 0.00 & 0.04 & \textbf{0.61} & \textbf{0.46}   \\ %
 644 & 0.00 & 0.03 & 0.00 & 0.01  & 0.00 & 0.04 & \textbf{0.76} & \textbf{0.75}   \\ %
 900 & 0.00 & 0.01 & 0.00 & 0.06  & 0.32 & 0.17 & \textbf{0.70} & \textbf{0.69}   \\ %
 949 & 0.00 & 0.01 & 0.00 & 0.02  & 0.00 & 0.06 & \textbf{0.77} & \textbf{0.79}   \\ %
 992 & 0.00 & 0.03 & 0.00 & 0.02  & 0.11 & 0.13 & \textbf{0.78} & \textbf{0.72}   \\ \bottomrule
                 
\end{tabular}
\end{table*}

\subsubsection{Porto Dataset}
\begin{figure}
     \centering
     \begin{subfigure}[b]{0.23\textwidth}
         \centering
         \includegraphics[width=\textwidth]{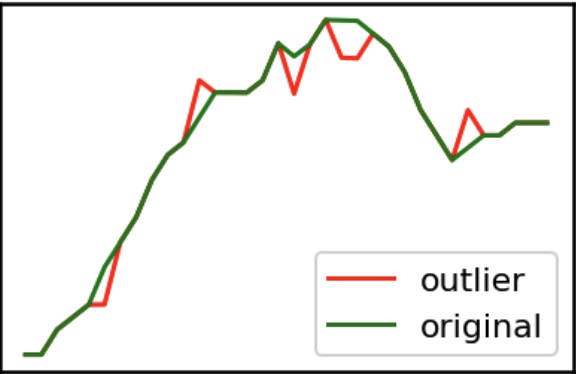}
         \caption{Random shift anomalies}
         \label{fig:random_shift_outlier}
     \end{subfigure}
     \hfill
     \begin{subfigure}[b]{0.23\textwidth}
         \centering
         \includegraphics[width=\textwidth]{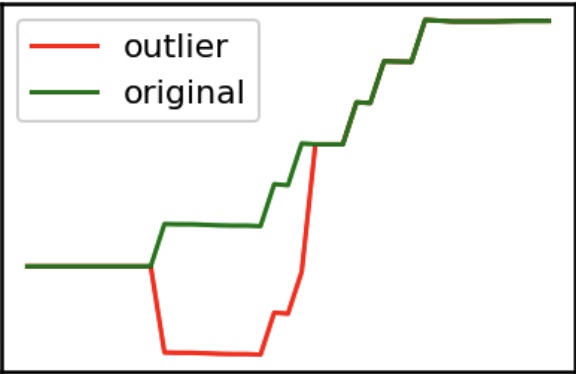}
         \caption{Detour anomalies}
         \label{fig:detour_outlier}
     \end{subfigure}
        \caption{Example of generated anomalies with $\alpha=0.3$ and $\beta = 3$ for both types of anomalies on the Porto dataset.}
        \label{fig:porto_outliers}
\end{figure}

The Porto dataset consists of data generated by 442 taxis operating in the city of Porto in Portugal from January 07, 2013, to June 30, 2014. A taxi reports its GPS location at 15s intervals. We employed preprocessing steps similar to \cite{LiuOnlineAnomalous2020} and \cite{Li2018DeepRepLearningforTraj}. We discretized the map into $100m \times 100m$ grids and group trajectories with the same source and destination. We discarded trajectories belonging to a ``source-destination'' group with fewer than 25 trajectories. The input to our model consists of a vector of chronologically ordered and discretized GPS coordinates (grid cells) prepended with SOT (start of trajectory) and appended EOT (end of trajectory) tokens.

Since this dataset did not have ground-truth labels for anomalous trajectories, we artificially generated anomalies following the work in \cite{LiuOnlineAnomalous2020} and \cite{ZhengContextualSpatial2027}. We created two types of anomalies, (i) random shift and (ii) detour anomalies. Figure~\ref{fig:porto_outliers} shows an example of the two types of outliers. For random shift anomalies, we perturb the $\alpha$ percentage of locations in a trajectory and move those location $\beta$ grid cells away. For detour anomalies, we create a detour for the $\alpha$ percentage portion of a trajectory and shift the detour $\beta$ grid cells away from the original trajectory. Following previous work on this dataset for anomaly detection \cite{Li2018DeepRepLearningforTraj}, we did not include the artificially generated anomalous data during training.

\subsubsection{Tokenization \& Vocabulary}
Given the nature of a language model architecture, we created tokens to form our model's vocabulary. In the Porto dataset, a token is considered a discretized GPS coordinate. We also added three special tokens: SOT (start of trajectory), EOT (end of trajectory), and PAD (padding token to help with batch training). In the POL dataset, tokens consist of staypoints (work, apartment, restaurant, etc.), days of the week (Monday, Tuesday, etc.), agent ID, and the EOT and PAD special tokens.

\vspace{-1em}
\subsection{Baselines}
We compared our method to existing unsupervised anomaly detection methods on trajectory data. Given the established better 
performance of deep learning methods on trajectory anomaly detection \cite{Li2018DeepRepLearningforTraj}, we omitted the inclusion of traditional clustering-based algorithms. 

\begin{itemize}
    \item \textbf{SAE}: a standard autoencoder method trained to optimize the reconstruction loss of a trajectory sequence using a recurrent neural network. Based on the work in \cite{Malhotra2016LSTMbasedEF} and \cite{An2015VariationalAB}, we use the reconstruction error as the anomaly score. 
    \item \textbf{VSAE} A method similar to SAE, however, in addition to optimizing for the reconstruction loss, it also optimizes the KL divergence between the learned distribution over the latent space and a predefined prior \cite{An2015VariationalAB, Sachdeva2018SequentialVA}. Similar to SAE, we use the reconstruction error as the anomaly score.
    \item \textbf{GM-VSAE} \cite{LiuOnlineAnomalous2020}. This method generalizes VSAE by modeling the latent space with more than one Gaussian component and also uses the reconstruction error as the anomaly score. 
\end{itemize}

\subsection{Evaluation Metrics}
We use Precision-Recall AUC and F1 scores to evaluate the performance of our method and the baseline methods \cite{LiuOnlineAnomalous2020, Zhang2023ATO}. These metrics are suitable for assessing the performance of anomaly detection methods as the number of anomalies in each dataset is small compared to normal trajectories. For the Porto dataset, these metrics are computed across all trajectories. Conversely, we conduct these evaluations on a per-virtual-agent basis for the Pattern-of-Life (PoL) dataset. In addition, the surprisal rate metric is used to locate the specific occurrence of an anomaly within a trajectory.

\section{Results}

The following sections present the anomaly detection results for the PoL and Porto datasets.
Anomaly detection for the Porto dataset is a \textit{global challenge}, since taxi movements are customer/ride-driven and the movements captured by individual trajectories are largely independent. 
This is in contrast to the PoL dataset, which contains sets of trajectories that model the behavior of individual agents, and anomaly detection will be agent-specific.

\begin{figure*}
     \centering
     \begin{subfigure}[b]{0.40\textwidth}
         \centering
         \includegraphics[width=\textwidth]{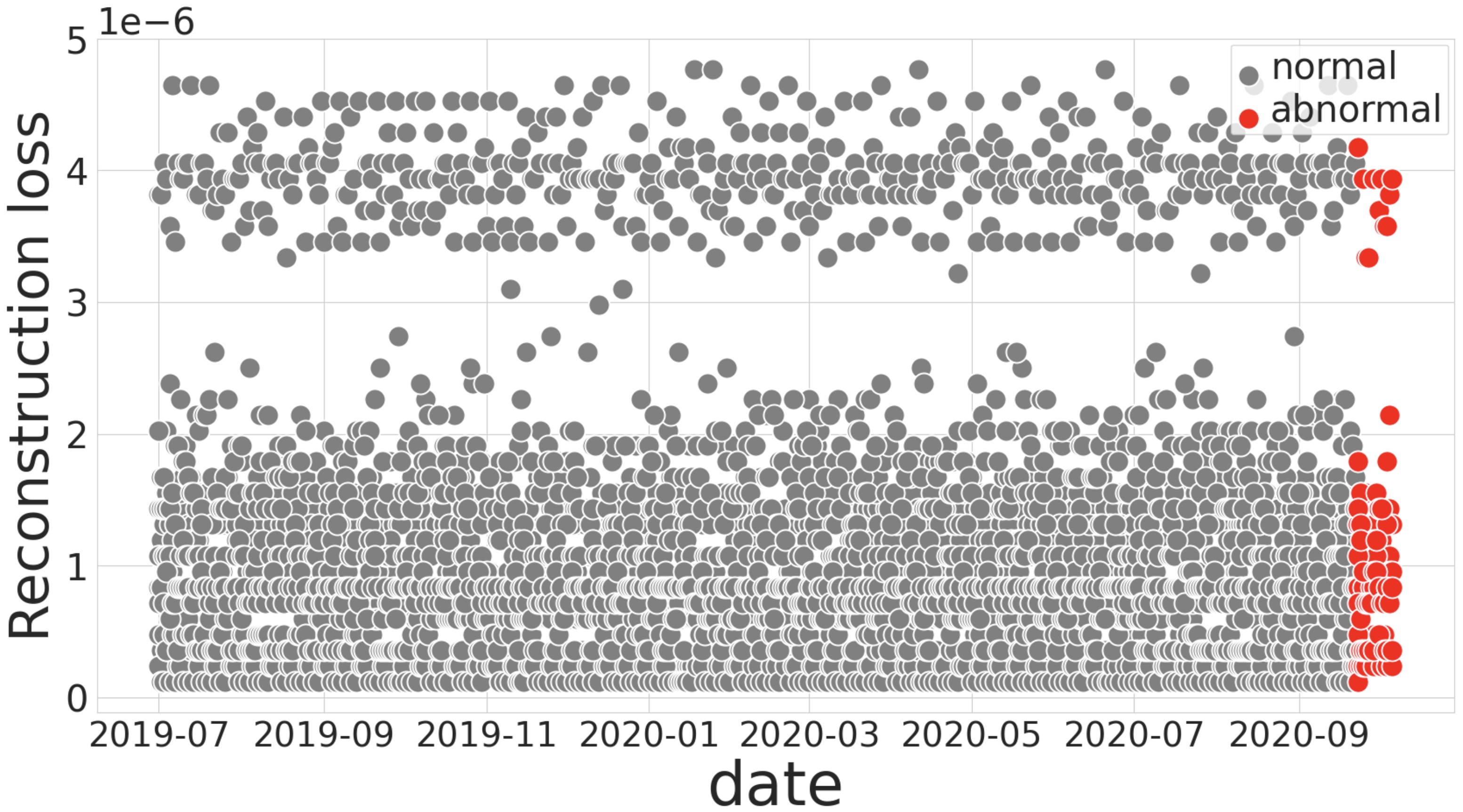}
         \caption{SAE}
         \label{fig:random_shift_outlier}
     \end{subfigure}
     \begin{subfigure}[b]{0.40\textwidth}
         \centering
         \includegraphics[width=\textwidth]{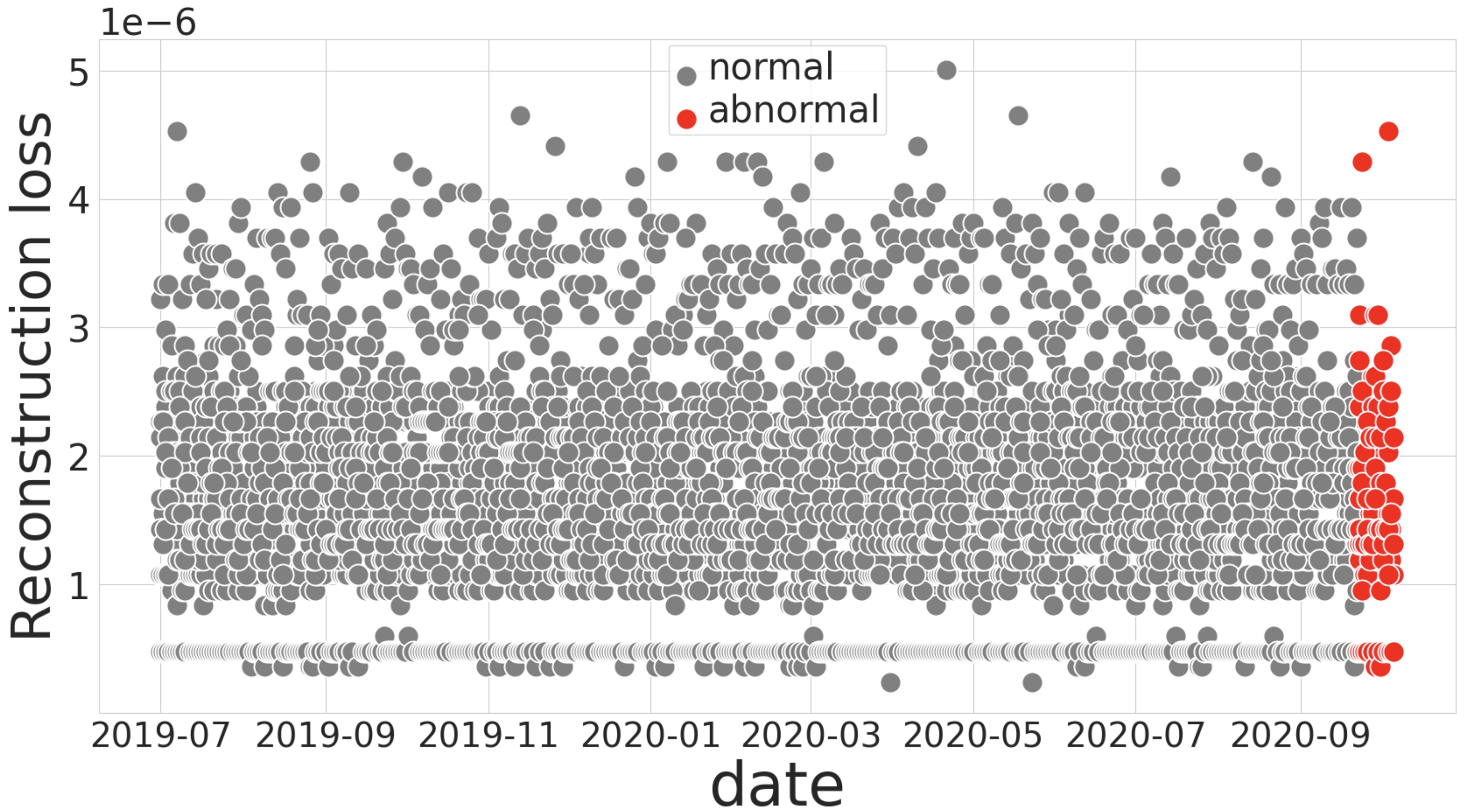}
         \caption{SVAE}
         \label{fig:detour_outlier}
     \end{subfigure}
     \hspace{1em}
     \begin{subfigure}[b]{0.40\textwidth}
         \centering
         \includegraphics[width=\textwidth]{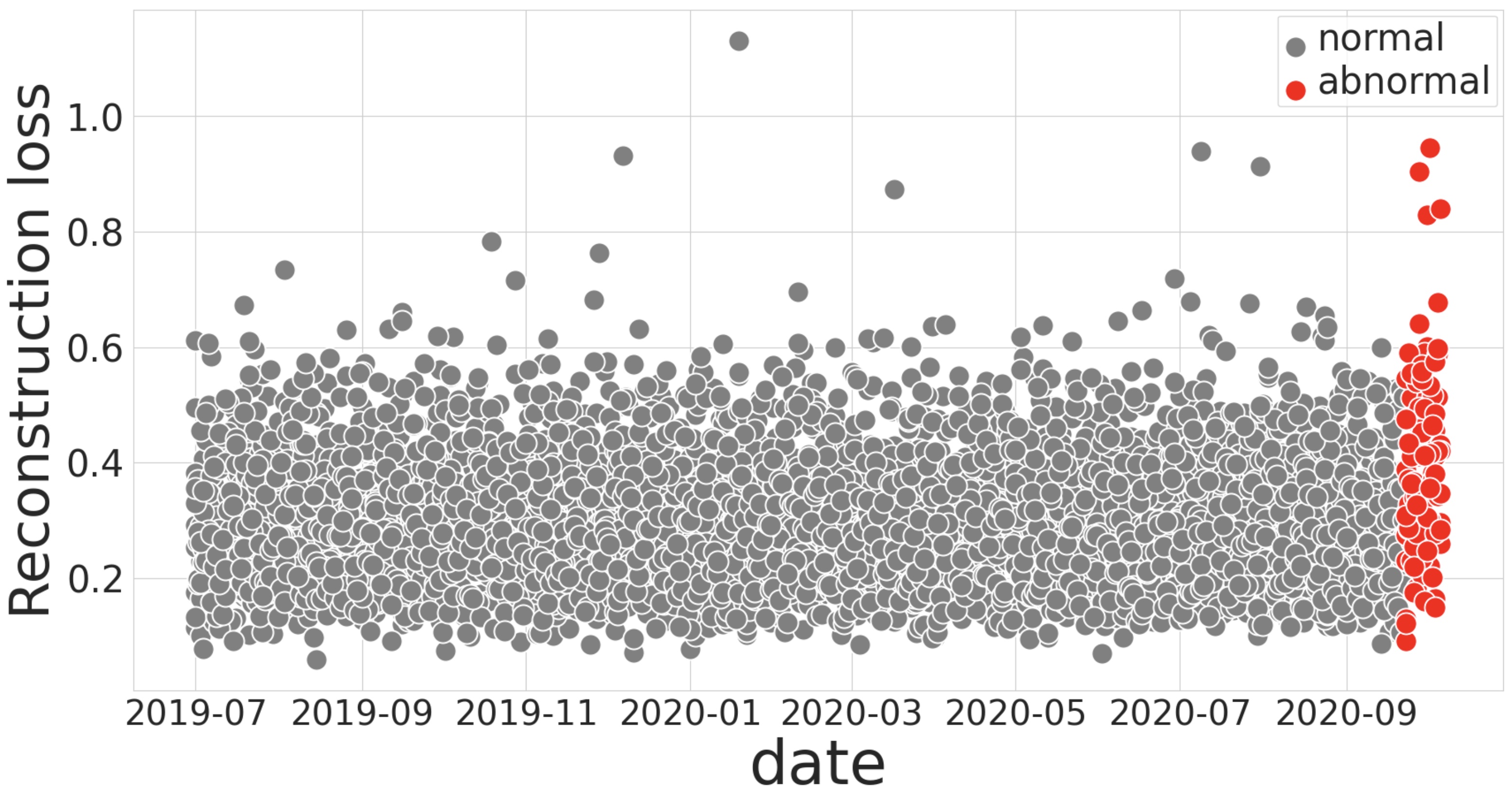}
         \caption{GM-SVAE}
         \label{fig:detour_outlier}
     \end{subfigure}
     \begin{subfigure}[b]{0.40\textwidth}
         \centering
         \includegraphics[width=\textwidth]{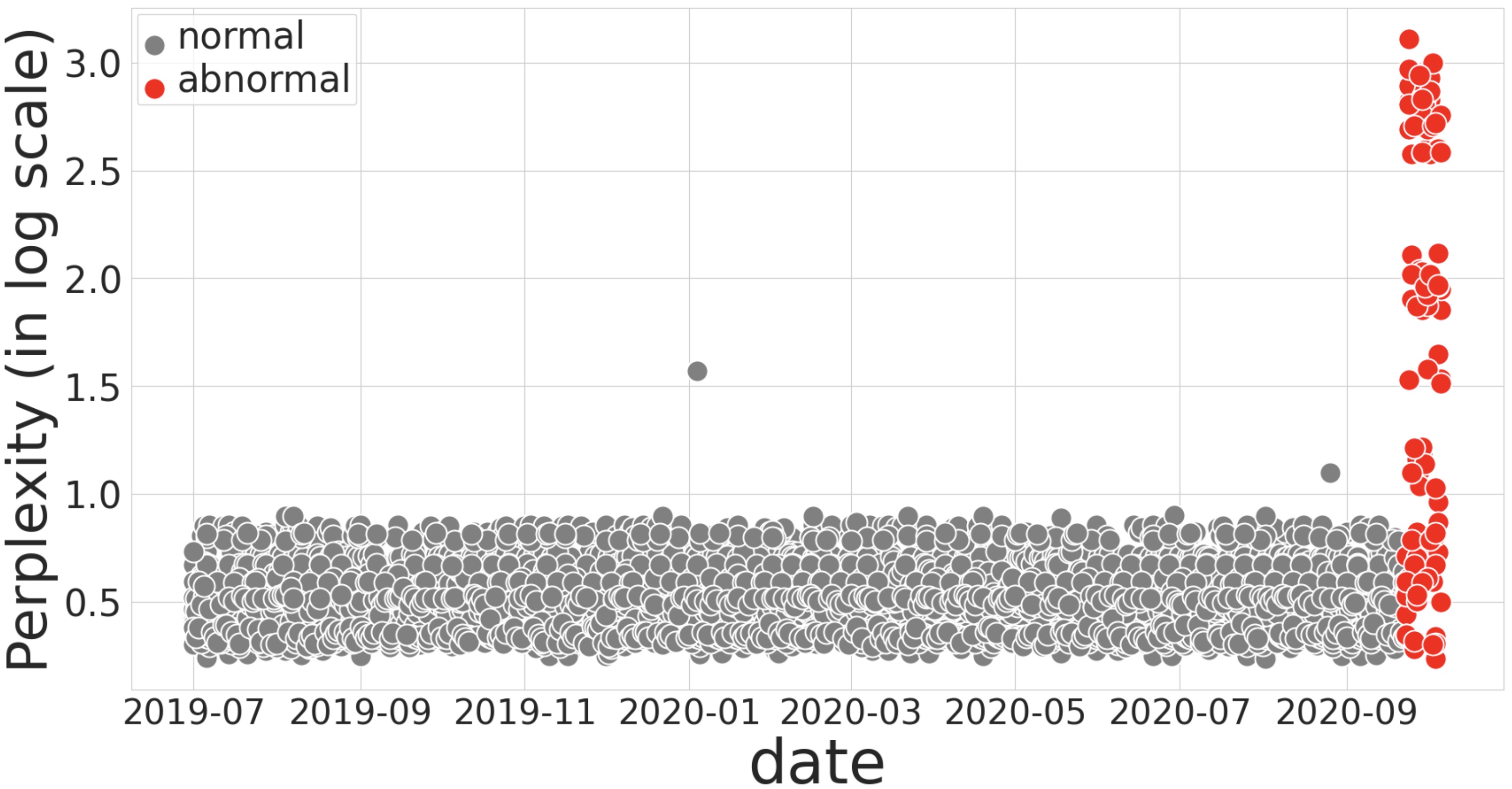}
         \caption{LM-TAD}
         \label{fig:detour_outlier}
     \end{subfigure}
        \caption{Anomaly results for all methods trained on all the pattern of life dataset trajectories. Each dot represents the perplexity of a trajectory for any of the ten agents with normal and anomalous trajectories. Unlike other methods, our method (d) distinguishes between anomalous trajectories and normal trajectories by scoring most anomalous trajectories with high perplexity. }
        \label{fig:porto_outliers}

    \label{fig:method_comparisions}
    
\end{figure*}

\subsection{Agent-based Outliers - Patterns-of-Life Data}
\label{pol_description}
The PoL dataset has user IDs, i.e., sets of trajectories that can be linked to a specific user. The anomaly detection challenge can become user-specific, as the anomalous behavior of one agent may be the normal behavior of another agent. Therefore, we report results for the ten virtual agents that each have 14 additional days of anomalous behavior. Table~\ref{tab:p_of_life_results} summarizes F1 and Precision-Recall Area-under-the-Curve (PR-AUC) results for these ten anomalous agents. The results of the rest of the agents are not included in table~\ref{tab:p_of_life_results} since these agents are not anomalous within their respective contexts.
\textit{\model{} outperforms all competitor methods} as it is more efficient in finding anomalies with respect to anomalous users. We identify three reasons why autoencoder approaches do not identify anomalies for the PoL dataset. 1) Autoencoder approaches are optimized to reconstruct the input by minimizing the reconstruction error. This training approach would tend to learn a model that overfits training data, which may contain anomalies and hence fail to distinguish normal and anomalous trajectories. 2) Autoencoder-based methods yield a high reconstruction error for inputs divergent from the overall data patterns within the dataset. Nonetheless, these methods are suboptimal for identifying anomalies on an individual-agent basis. Theoretically, GM-SVAE can model distinct Gaussian distributions that can correspond to each virtual agent, thereby learning trajectory distributions unique to each agent. However, this approach has little control over the distributions learned by each component in the latent dimension. Even if we had control over the distributions in the latent space, this approach would be very expensive to train in practice, as the increase in the number of agents would require an increase in Gaussian distributions. 3) The final reason is related to defining what we consider anomalies. Recent literature on trajectory anomaly data \cite{LiuOnlineAnomalous2020, Zhang2023OnlineSubtrajectoryDetection} considers normality as a trajectory that the majority of taxis in the train data took. Therefore, in this context, an anomalous trajectory was one that was not or rarely used by taxis. However, the PoL dataset's anomalies differ from the those of Porto. Anomalies in the PoL dataset deviate from their typical behavior, e.g., not going to work on days one should. Autoencoder approaches fail to capture these types of anomalies by not being able to learn the anomalous behavior on a per-agent basis. 

\model{} overcomes the limitations of the autoencoder approaches as follows. 1) \model{} can learn the likelihood that a particular agent will visit a particular location. Therefore, if an agent hardly visits a location, even if it was part of the training data, the model will give a low probability of such a location and distinguish anomalies. 2) \model{} uses a special token to provide the context for generating an agent's trajectory, therefore being able to find anomalies on an agent basis. 3) \model{} is highly customized for different types of anomalies. It can learn the normal pattern for each agent for each day of the week because the model can condition the generation of a particular trajectory to specific tokens. Therefore, this approach is still practically feasible even if the number of agents in the data increases. 

Moreover, Figure~\ref{fig:method_comparisions} illustrates why competitor methods have low F1 and PR-AUC scores. In this figure, the red dots represent the perplexity or reconstruction error associated with the agent trajectories. The expectation is that trajectories deemed anomalous (indicated by red dots) would yield higher levels of perplexity or reconstruction error. However, for autoencoder-based methods, we find that the reconstruction error associated with anomalous trajectories is comparatively lower than that of many normal trajectories.
In contrast, \model{} aligns with the expected model behavior and \textit{attributes higher perplexity scores to anomalous trajectories. }

\begin{table*}[t]
\caption{Anomaly detection results on the Porto dataset. The best results for a particular metric in a specific category are \textbf{bolded}. \model $\space$ performs largely on par with the best baseline.}
\label{tab:porto_results}
\centering
\begin{tabular}{r|cc|cc|cc|cc|cc|cc} \toprule
                   
& \multicolumn{6}{c|}{Random shift anomalies}    &  \multicolumn{6}{c}{Detour anomalies} \\ 

anomalies params:  &   \multicolumn{2}{c|}{$\alpha = 3$, $\beta = 0.1$} &   \multicolumn{2}{c|}{$\alpha = 5$, $\beta = 0.1$} & \multicolumn{2}{c|}{$\alpha = 3$, $\beta = 0.3$} &   \multicolumn{2}{c|}{$\alpha = 3$, $\beta = 0.1$} &   \multicolumn{2}{c|}{$\alpha =5$, $\beta = 0.1$} &   \multicolumn{2}{c}{$\alpha = 3$, $\beta = 0.3$} \\ \midrule
 
Metric & F1 & PR-AUC & F1 & PR-AUC  & F1 & PR-AUC & F1 & PR-AUC & F1 & PR-AUC & F1 & PR-AUC  \\ \midrule
\multicolumn{2}{l}{\ baselines:} \\[-.5em]
SAE & 0.44 & 0.66 & 0.53 & 0.75  & 0.71 & 0.90 & 0.30 & 0.47 & 0.36 & 0.54 & 0.58 & 0.78  \\ 
VSAE & 0.40 & 0.67 & 0.51 & 0.76  & 0.69 & 0.90 & 0.28 & 0.47 & 0.35 & 0.56 & 0.57 & 0.77  \\ 
GM-VSAE-10 & \textbf{0.85} & 0.90 & \textbf{0.86} & \textbf{0.91}  & \textbf{0.91} & \textbf{0.99} & \textbf{0.73} & \textbf{0.72} & \textbf{0.79} & \textbf{0.77} & \textbf{0.90} &\textbf{ 0.96}  \\ \midrule
ours: LM-TAD & \textbf{0.85} & \textbf{0.91} & 0.85 & \textbf{0.91}  & 0.90 & \textbf{0.99} & 0.69 & 0.65 & 0.73 & 0.68 & 0.89 & \textbf{0.96}  \\ \bottomrule
                 
\end{tabular}
\end{table*}

\subsection{Global Outliers - Porto Taxi Data}

The results for the Porto taxi dataset are summarized in Table~\ref{tab:porto_results}. 
We use different parameter configurations $\alpha$ and $\beta$ for random-shift and detour anomalies. For random shift, we perturb $\alpha$ percent of locations in a trajectory, moving them $\beta$ grid cells away. Similarly, for detour anomalies, we create a detour for $\alpha$ percent of a trajectory, shifting the detour $\beta$ grid cells away from the original trajectory. 

Our method outperforms the SAE and VSAE methods. For random shift anomalies, \model{} shows performance comparable to that of GM-SVAE. For detour anomalies, GM-SVAE is able to identify more anomalies than our method, especially for cases where the detour is about 10\% of the entire trajectory. 
This behavior of \model{} can be explained by the perplexity of being less sensitive to capturing anomalies that consist of changing a small continuous portion of the trajectory (detour). Since most continuous locations in the trajectory are normal, and as such, those probabilities are fairly high, the overall perplexity will also be relatively high. Conversely, for random shift anomalies, our method exhibits comparable and often superior performance to GM-SVAE. This enhanced detection efficacy can be attributed to the fact that the random shift anomalies break the continuity dependence of one location to its history, resulting in a sequence of locations with lower probabilities and consequently a lower perplexity score.

We can also observe that the distance ($\beta$) of the detour or the randomly shifted location from the original trajectory impacts finding anomalous trajectories less than the anomalous trajectory fraction ($\alpha$). This suggests that metrics that tend to summarize the anomaly of a trajectory by looking at the entire trajectory (i.e., reconstruction error) may find identifying anomalous trajectories with few anomalous locations challenging. Consequently, we introduce a local \textit{surprisal rate} metric for such cases to address this limitation.

\subsection{Identifying Anomalies using Surprisal Rate}

Perplexity as an aggregate measure may not be sufficient to identify anomalous trajectories. 
Consequently, the presence of only a few anomalous tokens may lead to their signal being diluted by the averaging process and anomalies go undetected. Similar limitations apply to autoencoder-based methods, where the reconstruction loss is calculated over all tokens in a trajectory. 

A further limitation in using perplexity or reconstruction error is the inability to \textit{pinpoint the specific location of anomalies within a trajectory}. Here, our work proposes the \textit{surprisal rate}  measure that operates at the level of individual locations or tokens within a trajectory.

\begin{figure*}
\includegraphics[width=0.9 \textwidth]{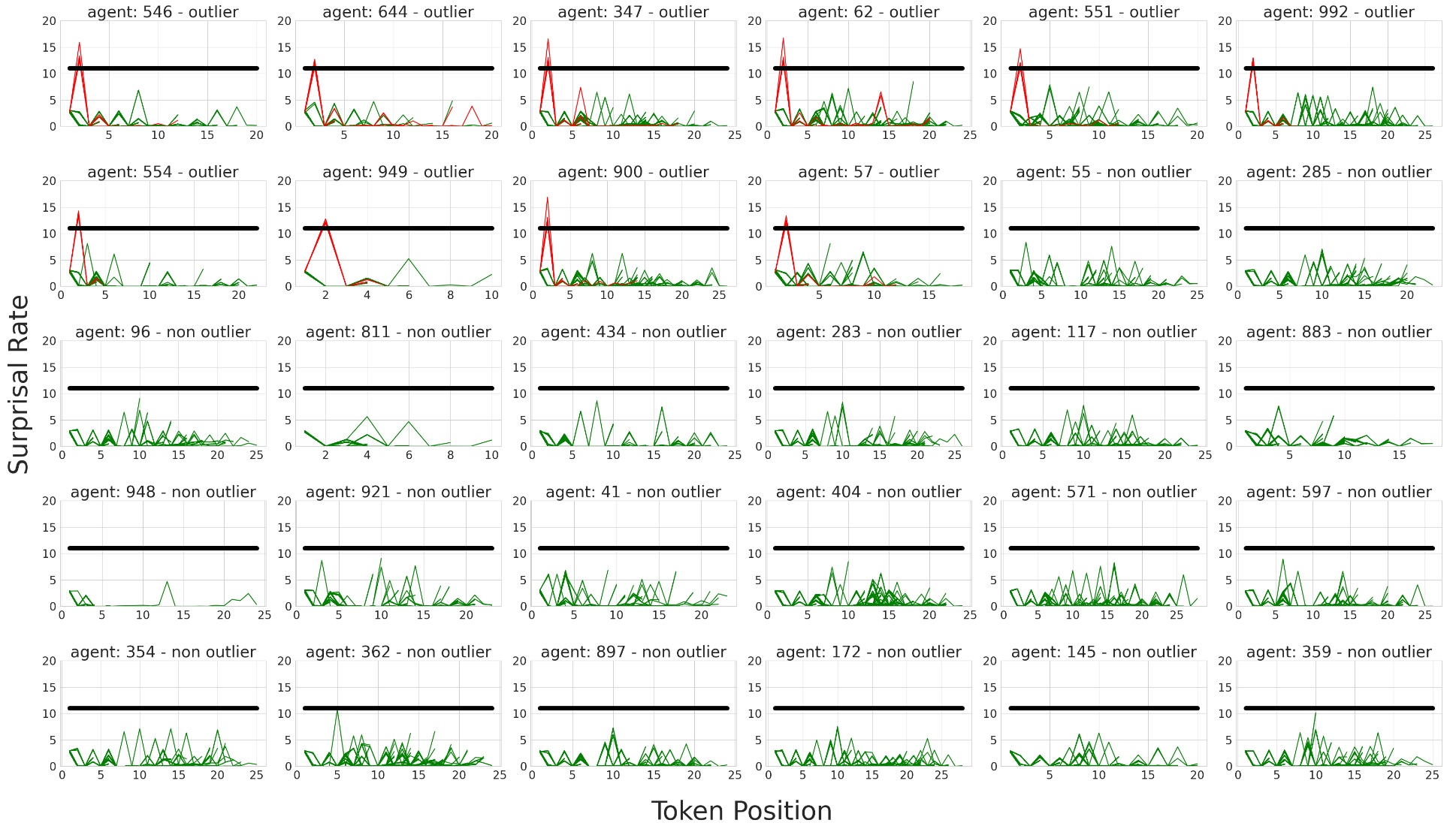}
\caption{Surprisal rate through trajectories on the Pattern of Life dataset. We plot the trajectories of each agent with some anomalous trajectories (10 first agents), and we randomly selected 20 agents with no anomalous trajectories. The horizontal line shows the (arbitrary) threshold 
 for a surprisal rate high enough to correspond to anomalous trajectories. Anomalous trajectories plotted in red have high surprisal rates for certain locations in the trajectories as opposed to normal trajectories. Hence, \model{} can identify anomalous trajectories based on the surprisal rate.} 
\label{fig:surprisal_rate_pattern_of_life}
\end{figure*}

In our empirical analysis, we explored the application of the surprisal rate for detecting potentially anomalous locations within a trajectory in the PoL data. A high surprisal rate suggests that a particular location in a trajectory may be anomalous. Figure~\ref{fig:surprisal_rate_pattern_of_life} shows the surprisal rate for 30 trajectories (10 anomalous and 20 normal ones randomly chosen from the respective agents). The analysis reveals that certain tokens in anomalous trajectories exhibit significantly higher surprisal rates compared to those in normal trajectories, particularly at the beginning of the trajectories. 
This pattern aligns with the dataset's structure and the configuration of our input vector, where the initial tokens represent the agent ID, the weekday, and the first location visited by the virtual agent on that day. Given the expected pattern of agents visiting consistent locations on specific weekdays, deviations from this routine, such as visiting an atypical location as the first destination, are flagged as anomalies. Consequently, the inclusion of the weekday token in the trajectory analysis enables the identification of instances where an agent's initial location deviates from the norm, resulting in a larger surprisal rate when an agent visits an unusual place on a given weekday.

\subsection{Location Configurations - Ablation Study}
We conducted an ablation study to show the versatility of \model{} in working with different types of inputs. In this study, we explore the usage of discretized GPS coordinates (Uber hexagons \cite{uberh3}), staypoint labels (i.e., work, restaurant, and so forth), and stay duration (the duration at a particular location) as input for the model to infer the anomaly detection performance of each modality. One of the main advantages of using one modality over another is the type of anomalies we are interested in discovering. Anomalies can be related to the duration of stay in a particular location (staying longer than usual), by visiting an uncommon geographical area, or by visiting a different place (e.g., shopping mall) on a weekday when one is supposed to be elsewhere (e.g., work).

Table~\ref{tab:location_configuations} summarizes the results of using various location configurations in the PoL dataset. Staypoint labels provide the best performance to identify anomalies. This is consistent with the anomalies in the PoL dataset as discussed in Section~\ref{pol_description}, where agents abstain from visiting places they would otherwise visit on certain days. %
The dwell time also proves to be effective since visiting different locations affects the time spent at those locations. These findings underscore the adaptability of our approach to using different feature configurations to identify anomalies.

\begin{table}[t]
\caption{Anomaly detection using different location configurations. The best results are highlighted in bold. The staypoint label location type performs best overall, reflecting the types of anomalies present in the PoL dataset.}
\label{tab:location_configuations}
\centering
\begin{tabular}{r|cc|cc|cc} \toprule
                   
  & \multicolumn{2}{c|}{Staypoint label} &   \multicolumn{2}{c|}{Discretized GPS} & \multicolumn{2}{c}{Stay duration}  \\ 
 
Agent & F1 & PR-AUC & F1 & PR-AUC  & F1 & PR-AUC    \\ \midrule
 57 & \textbf{0.62} & \textbf{0.69} & 0.50 & 0.54  & 0.26 & 0.49   \\ %
 62 & 0.86 & \textbf{0.81} & \textbf{0.88} & 0.80  & 0.78 & 0.80  \\ %
 347 & \textbf{0.78} & \textbf{0.75} & 0.75 & 0.68  & 0.72 & 0.70    \\ %
 546 & \textbf{0.75} & \textbf{0.76} & 0.67 & 0.65  & 0.67 & 0.62    \\ %
 551 & 0.67 & \textbf{0.63} & \textbf{0.76} & 0.63  & 0.52 & 0.48   \\ %
 554 & \textbf{0.62} & 0.46 & 0.53 & \textbf{0.61}  & 0.62 & 0.46    \\ %
 644 & \textbf{0.80} & \textbf{0.78} & 0.57 & 0.70  & 0.64 & 0.64    \\ %
 900 & \textbf{0.71} & \textbf{0.66} & 0.46 & 0.60  & \textbf{0.71} & 0.59  \\ %
 949 & \textbf{0.82} & \textbf{0.77} & 0.78 & 0.75  & 0.82 & 0.73    \\ %
 992 & \textbf{0.78} & 0.68 & 0.72 & \textbf{0.73}  & 0.67 & 0.69    \\ 
 \midrule
average & \textbf{0.74} & \textbf{0.70} & 0.66 & 0.67 & 0.64 & 0.62 \\
 \bottomrule         
\end{tabular}
\end{table}

\begin{figure*}[t]
    \centering
    \begin{subfigure}[b]{0.34\textwidth}
        \centering
        \includegraphics[width=\textwidth]{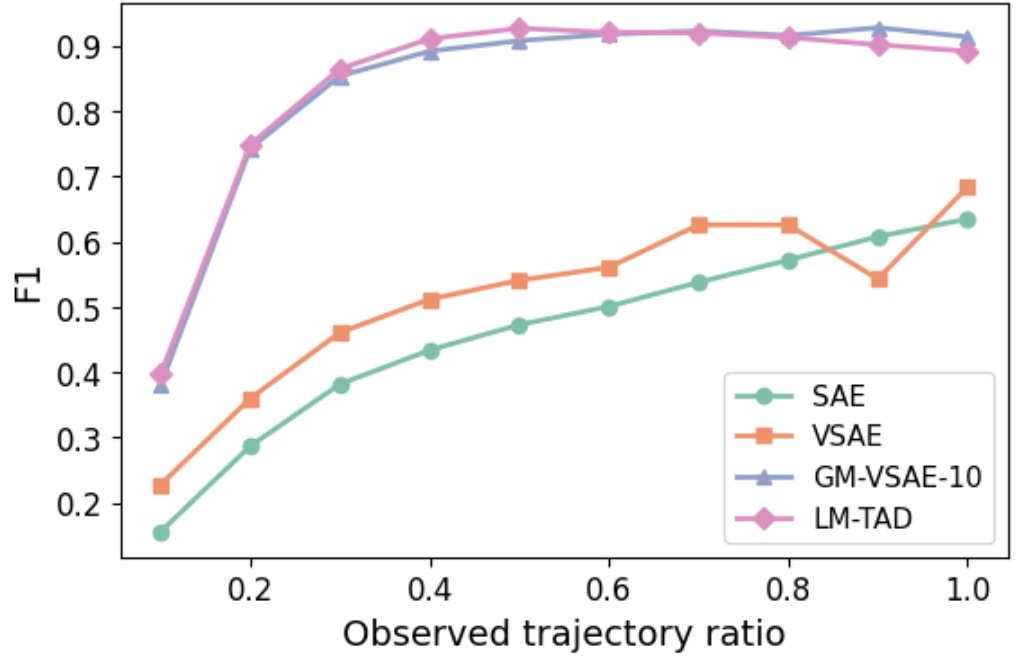}  %
        \caption{F1 scores for random shift anomalies}
        \label{fig:sub1}
    \end{subfigure}
    \begin{subfigure}[b]{0.34\textwidth}
        \centering
        \includegraphics[width=\textwidth]{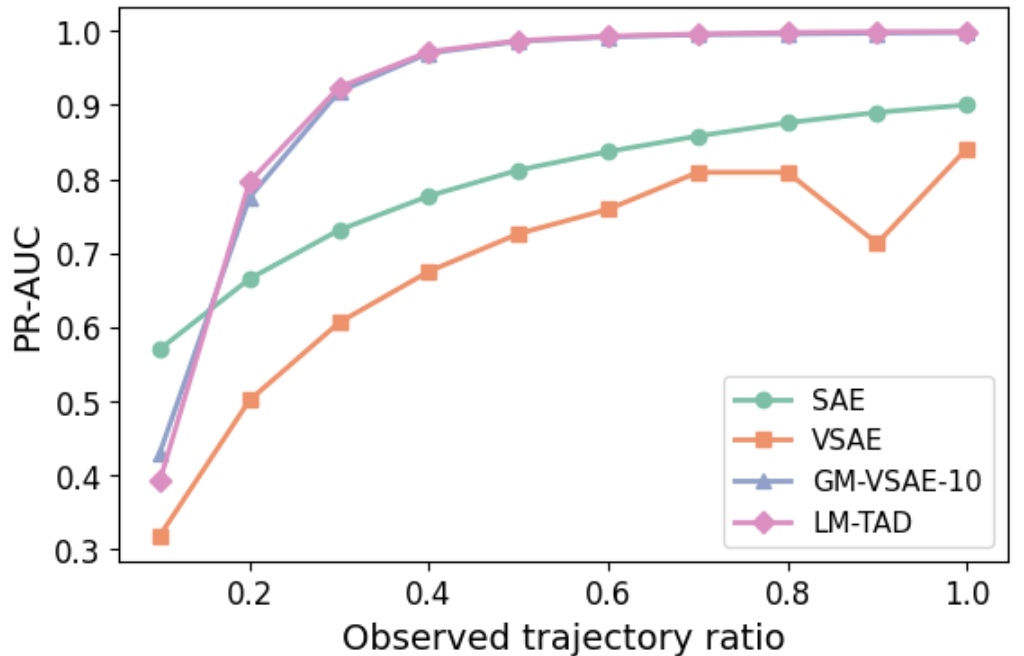}  %
        \caption{PR-AUC scores for random shift anomalies}
        \label{fig:sub2}
    \end{subfigure}
    \begin{subfigure}[b]{0.34\textwidth}
        \centering
        \includegraphics[width=\textwidth]{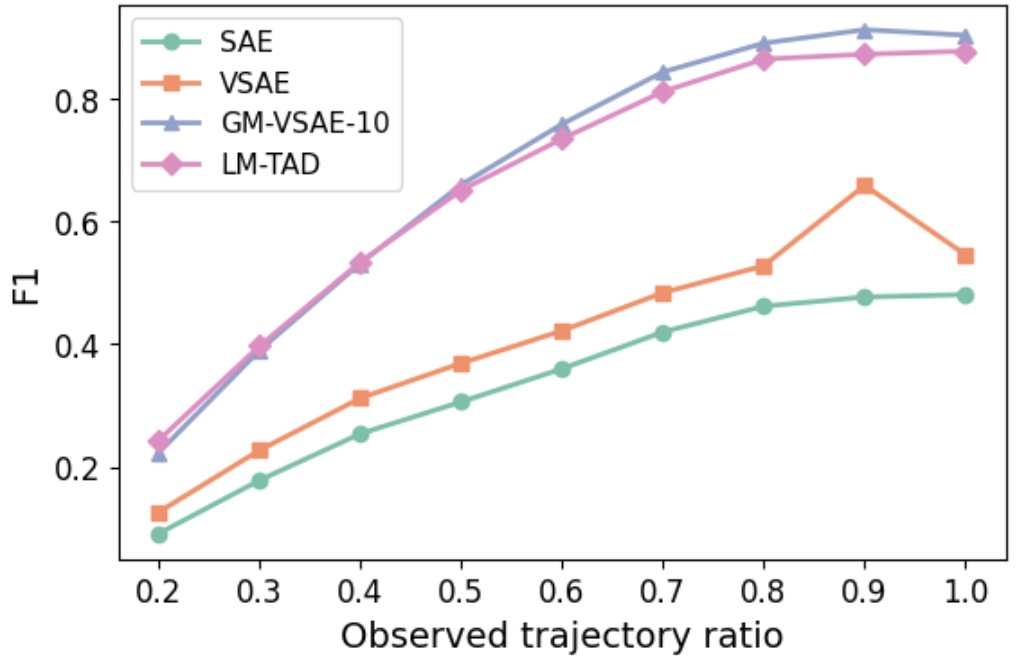}  %
        \caption{F1 scores for detour anomalies}
        \label{fig:sub3}
    \end{subfigure}
    \begin{subfigure}[b]{0.34\textwidth}
        \centering
        \includegraphics[width=\textwidth]{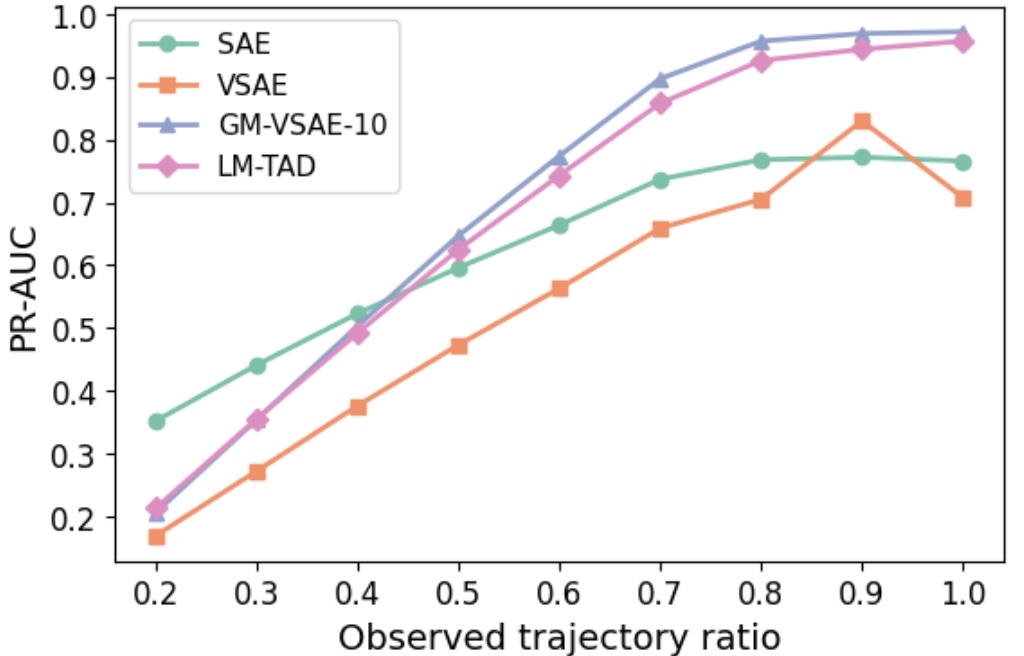}  %
        \caption{PR-AUC scores for detour anomalies}
        \label{fig:sub4}
    \end{subfigure}
    \caption{Online Anomalous Trajectory Detection Results (POL data). The results show the performance of each model for the detour and random shift anomalies as we evaluate different ratios of the trajectory from 0.1 to 1.0. \model{} shows competitive results and can detect anomalies of sub-trajectories. }
    \label{fig:online_detection}
    \vspace{-1em}
\end{figure*}

\subsection{Online Anomaly Detection}

One of the advantages of \model{} is the support of online anomalous trajectory detection. Our approach does not require the entire trajectory to compute an anomaly score. In addition, we do not need to know the destination (although such knowledge would enhance the anomaly detection of sub-trajectories). As soon as a trip begins, \model{} can compute the anomaly score of a partial trajectory each time a new location is sampled. 
Autoencoder approaches can be used for online anomalous trajectory detection as well. However, they are significantly more expensive to use since they must compute the anomaly score for the entire sub-trajectory each time a new location is sampled. \model{}, instead, can cache the key-value (KV cache) states \cite{Liu2024MiniCacheKC, Pope2022EfficientlyST} of the attention mechanism for previously generated tokens (i.e., GPS coordinates). This significantly reduces the need for repetitive computations and lowers the latency in computing the anomaly score.

Figure~\ref{fig:online_detection} shows the accuracy of detecting anomalies for partial trajectories at different completion ratios for the Porto dataset. We evaluate partial trajectories with ratios from 0.2 to 1.0 (complete) using 0.1 increment. The results suggest that \model{} is more than competitive in detecting sub-trajectory anomalies. Especially for random shift anomalies, 40\% of the sub-trajectory is sufficient to detect most anomalies in the dataset. Detour anomalies are not that easily detected for small completion ratios without knowing the destination. A detour becomes  evidently anomalous only when knowing the destination or a large portion of the trajectory. In general, \model{} performs on par with the best baseline, but as discussed before, comes with the advantage of significantly lower latency.

\section{Conclusions} \label{conclusion}

In this work, we introduced \model{}, an innovative trajectory anomaly detection model that uses an autoregressive causal-attention mechanism. By conceptualizing trajectories as sequences akin to language statements, our model effectively captures the sequential dependencies and contextual nuances necessary for precise anomaly detection. We demonstrated that incorporating user-specific tokens enhances the model’s ability to detect context-specific anomalies, addressing the variability in individual behavior patterns.

Our extensive experiments validated the robustness and adaptability of \model{} across two datasets, the Agent-based-Model generated Pattern-of-Life (PoL) dataset and the Porto taxi dataset. The results show that \model{} vastly outperforms existing state-of-the-art methods in identifying user-contextual anomalies. At the same time, its performance is competitive in detecting outliers in GPS-based trajectory data.

We introduced perplexity and surprisal rate as metrics for outlier detection and localization of anomalies within trajectories, broadening the analytical capabilities of the approach. The model's ability to handle diverse trajectory representations, from GPS coordinates to staypoints and activity types, underscores its versatility and uniqueness.

Importantly, our approach also proves advantageous for online trajectory anomaly detection, reducing computational latency, and gaining a significant performance advantage over existing models. This provides for real-time anomaly detection without the need for an expensive re-computation of results.

In summary, \model{} represents a substantial advance in trajectory anomaly detection, offering a \textit{scalable}, \textit{context-aware}, and \textit{computationally efficient} solution. This work paves the way for future research in user-centric analysis and real-time anomaly detection in trajectory data.

\section*{Acknowledgment}

The authors thank Andreas Zuefle for making the Atlanta PoL dataset available. This work was supported by the National Science Foundation (Award 2127901) and by the Intelligence Advanced Research Projects Activity (IARPA) via Department of Interior/ Interior Business Center (DOI/IBC) contract number 140D0419C0050.  The U.S. Government is authorized to reproduce and distribute reprints for Governmental purposes, notwithstanding any copyright annotation thereon. Disclaimer: The views and conclusions contained herein are those of the authors and should not be interpreted as necessarily representing the official policies or endorsements, either expressed or implied, of IARPA, DOI/IBC, or the U.S. Government. Additionally, this work was supported by resources provided by the Office of Research Computing, George Mason University and by the National Science Foundation (Award Numbers 1625039, 2018631). 

\newpage
\bibliographystyle{ACM-Reference-Format}
\bibliography{main}

\clearpage
\appendix

\end{document}